# Federation Is Nearly Free, Reasoning Is Not: Tradeoffs for AI Co-Scientists in Protein Characterization Workflows

Maia S. Kapur[1], Timothy Boe[1], Abby Jerger[1], Paul Rigor[2]
[maia.kapur]ATpnnl.gov
ORCiD: 0000-0002-1781-169X

**Abstract**

Natural language driven autonomous co-scientist workflows involve a fundamental trade-off between flexibility and reasoning at the expense of determinism, reproducibility, and observability. Such agents increasingly must communicate across institutional boundaries, where federation topology can shape latency and cost. We systematically evaluated these tradeoffs using a controlled ablation on a production agentic platform for science. We use a verifiable task: given a protein sequence, we ask an agent to confidently characterize its function by routing across common tools. We compare federation topology, classic RL vs LLM-driven harnesses, language model, and prompt expertise. We also stratify results by protein novelty. We find that the choice of LLM dominated prediction quality far more than topology or prompting (Opus ~92%-94% vs o4-mini ~40%-50%). The PPO policy was nearly as accurate as the best LLM (88%) at zero token cost, fastest latency, and perfect consistency, but yields no reasoning trace. Expert prompted LLMs reached the highest accuracy but were high-cost and less consistent; prompt dependence was largest when the task was hardest. Federation imposed a negligible penalty on performance. These results offer actionable guidance for deploying agents for scientific workflows: for routine, verifiable tasks, a cheap deterministic policy delivers near-frontier accuracy with complete reproducibility, while flexible LLM reasoning is best reserved for open-ended discovery.

## Introduction

The past two years have seen rapid proliferation of autonomous AI systems designed to accelerate scientific discovery, variously termed "co-scientists," "AI lab assistants," or "scientific agents". At their core, these systems share a common architecture: a large language model wrapped in an orchestration harness that provides access to domain-specific tools, literature retrieval, and/or iterative planning capabilities. The sophistication of these systems varies, from industry platforms like Google's AI Co-Scientist (Gottweis et al., 2026) and multi-facility deployments such as Agentic Discovery and Exploration Platform for Tools' (ADEPT, George et al., 2025) federated agent mesh across DOE national laboratories, through systems like ProtAgents (Ghafarollahi and Buehler, 2024), SciAgents (Ghafarollahi and Buehler, 2025), and STELLA (Jin et al., 2026), down to implementations for individual labs or teams built on frameworks like LangChain that connect a single large language model (LLM) to a data and tooling source. These approaches are unified by a definition of "agentic" that goes beyond single-turn question answering to decompose goals into sub-routines, tool invocations, and iterative, chain-of-thought based reasoning and refinement. This think-act-observe loop approximates the design-build-test-learn cycle familiar to laboratory scientists and occupies a pragmatic middle ground that augments rather than replaces human researchers. A comprehensive survey by Ren et al. (2026) describes the core architectural components (Planner, Memory, Action Space, and Verifier) across such co-scientist systems.

[1] Environmental Molecular Sciences Laboratory, Pacific Northwest National Laboratory, 3335 Innovation Blvd, Richland, WA, USA 99354
[2] Research Computing and Continuum Platform Engineering, Pacific Northwest National Laboratory, 902 Battelle Blvd, Richland, WA, USA 99352

Importantly, these systems' autonomy derives primarily from orchestration logic rather than learned policies. Most deployed co-scientists today do not employ state-of-the-art reinforcement learning (RL) methods like group relative policy optimization (GRPO, Shao et al., 2024) or the incorporation of verifiable rewards (Wen et al., 2025) for policy optimization; they instead rely on prompt engineering, few-shot examples, and tool-augmented generation within a largely frozen, third-party LLM, with orchestration handled by deterministic graph executors rather than learned policies. The trend toward hybrid planning architectures that combine prompt-based flexibility with RL-optimized components is acknowledged in the literature (Ren et al., 2026) but remains aspirational for most practical deployments.

The relationship between RL and LLM-based scientific agents is well-studied but almost exclusively framed as complementary rather than comparative. The survey (Ren et al., 2026) of LLM-based scientific agents explicitly categorizes planners into prompt-native and learned types, with learned planners that can include RL-based approaches. This posits RL as one design choice within an LLM agent's planning module rather than as an alternative paradigm. Examples like ReFT (PPO + chain-of-thought for mathematical reasoning, Luong et al., 2024) and SciMARL (multi-agent RL for simulation design, Bae and Koumoutsakos, 2022) are RL-augmented LLM systems, not parallel RL harnesses evaluated against LLM harnesses on identical tasks.

A parallel thread explores LLM-RL collaboration as mutual enhancement. ExploRLLM uses LLMs to provide exploration policies that guide RL agents in robotic manipulation, treating foundation model outputs as exploratory behaviors within an RL framework rather than as competing controllers (Ma et al., 2025). The bi-directional feedback framework proposed by Gu demonstrates how LLMs can provide abstract planning information to RL agents while RL agents provide real-time advantage-based feedback to improve LLM token generation a teacher-student cooperation rather than a paradigm comparison (Gu, 2025). Similarly, the reinforcement learning with tool rewards framework optimizes tool-use sequences *inside* LLM agents via reinforcement learning with tool-use completeness rewards, improving action planning performance by 8-12% (Li et al., 2025). In all cases, RL serves as a training signal or collaborative mechanism for the LLM agent's planning capability, not as an independent controller operating over the same environment. This distinction matters because when RL is embedded within the LLM loop, it optimizes how the LLM plans, but commits the user to the LLM-driven harness and associated costs. Our interest is whether a classical RL agent can accomplish the same scientific task as an LLM harness, and what the observable tradeoffs look like when both operate through shared infrastructure.

Scientific benchmarks for LLM agents evaluate performance on increasingly realistic tasks but universally omit classical RL/PPO agents as baselines. HeurekaBench (Panigrahi et al., 2026) introduces open-ended research questions grounded in real scientific workflows, evaluating co-scientists on exploratory data analysis, but only compares LLM-based agents against each other. ScienceAgentBench (Chen et al., 2025) compiles 102 tasks from peer-reviewed publications across four disciplines, finding that even the best-performing agent (Claude-3.5-Sonnet with self-debug) solves only 34.3% of tasks, and their experimental design also only considers LLM frameworks, not alternative control paradigms. SciAssess (Cai et al., 2024) evaluates LLMs across progressive cognitive levels from memorization through analysis and reasoning in scientific literature, again without non-LLM baselines. Conversely, RL benchmarks in robotics and control use PPO or GRPO extensively but do not include LLM-driven harnesses as alternative controllers operating over identical reward structures (Wang et al., 2025).

This results in a separation between experimental design paradigms for science-oriented, multi-turn tasks resembling the iterative workflows co-scientists must handle. The few comparative examples that exist have demonstrated that minimalist RL techniques can significantly enhance reasoning in models as small as 1.5B parameters (Dang and Ngo, 2026) focus on mathematical benchmarks or tool-use optimization, not scientific discovery workflows with real tool costs, API latencies, and federation overhead.

Given this landscape, our study occupies a specific and largely unoccupied niche. We do not propose novel RL algorithms; we use off-the-shelf PPO via stable-baselines3 (Raffin et al., 2021). We do not claim to advance LLM reasoning capabilities. Rather, we ask a pragmatic question relevant to anyone building a co-scientist system: *on a controlled scientific task (protein function characterization), how does a classical PPO controller compare to an LLM-driven harness when both operate through the same federated orchestration infrastructure, measured on unified axes of cost, accuracy, latency, and observability?*

This framing is deliberately modest in methodological novelty and scale but practically significant. We did not use pre-existing life science benchmarks because we wanted to make use of the existing toolset within the ADEPT framework, specifically given its anticipated role in the forthcoming Genesis mission (Exec. Order No. 14,363, 2025). As institutions invest in co-scientist platforms, particularly federated deployments spanning multiple facilities with heterogeneous tools and data sovereignty requirements, understanding which components benefit from learned policies versus LLM reasoning, and at what cost, becomes an engineering decision with real resource implications.

While recent surveys and benchmarks study RL-based planners within LLM scientific agents and evaluate LLMs on scientific tasks without RL baselines, we find no prior work that directly compares a classical PPO controller to an LLM-driven harness on the same scientific discovery environment with unified metrics for cost, speed, accuracy, and observability. Our work offers preliminary guidance for co-scientist developers navigating an increasingly complex design space, where the choice between paradigms, or combinations thereof, carries implications for budget, latency, interpretability, and scientific outcome quality.

## Methods

The study was designed to mimic an agent-driving autonomous life sciences workflow to facilitate scientific discovery for an individual or federated laboratory. The existing observability infrastructure within ADEPT (Langfuse tracing, per-tool cost attribution, multi-tier session instrumentation, and cross-gateway delegation telemetry) provides the instrumentation to make these tradeoffs legible (George et al., 2025); it is also being implemented for use in the American Science Cloud, for which inter-laboratory federation will be a core component. The platform's natural decision points (planner, supervisor, model router) already make sequential decisions that map directly to RL state-action formulations and its rich telemetry provides reward signals without additional instrumentation

To provide a verifiable benchmark for our investigations, we identified and downloaded target proteins from UniProt (The UniProt Consortium et al., 2025), BLAST with Swiss-Prot (Basic Local Alignment Search Tool, Camacho et al., 2009) and AlphaFold (Jumper et al., 2021) servers (details in the Data section below). During each trial, an agent was asked to predict protein function, defined by numeric Gene Ontology (GO) annotations, based on the accession ID and/or genetic sequence associated with the protein. We contextualize our results by the confidence level reported in BLAST. For our main results we ran the experiments against all test data five times, which enables us to compare across methods and calculate inter-protein prediction stability. The following sections detail the components of this workflow.

## Data

Our benchmark data was assembled from three public sources via their REST/web APIs. UniProt entries were fetched from the UniProtKB REST API (rest.uniprot.org), storing both the full entry JSON and the FASTA sequence per accession; for each query we additionally BLASTed its top-3 hits and cached those hit entries as JSON (uniprot_hits/). BLAST was run through NCBI's blastp web service against the SwissProt database with E-value cutoff 10.0 and a 5-hit list, storing truncated per-hit summaries (accession, title, bit score, E-value, % identity, alignment coordinates) as JSON rather than full alignments; in this study, "BLAST confidence" is the reported percent identity (identical residues/alignment length), scaled between 0 and 1. AlphaFold structure predictions were pulled from the EBI AlphaFold API (alphafold.ebi.ac.uk/api/prediction), caching confidence metrics from the providers (pLDDT or predicted local distance difference test), model version, and file URLs as JSON. Structural files themselves were referenced by URL, not downloaded.

We selected organisms to ensure broad coverage across BLAST confidence regimes and evolutionary divergence, to create a dataset where the RL agent faced genuinely varied decision contexts (Figure 2). The high-confidence regime included well-characterized bacterial proteomes (*E. coli*, *P. putida*) that are densely represented and return high-identity matches with BLAST, making local sequence lookup the rational agent option. To populate the low-confidence regime (where BLAST identity falls below 70% and delegation to AlphaFold becomes the rational choice) we added two deeply divergent eukaryotes: *N. gruberi*, a free-living amoeba with minimal Swiss-Prot homologs, and *H. vulgaris*, an early-branching cnidarian whose proteome contains many novel domain architectures poorly represented in curated databases. We specifically targeted entries from both organisms with annotation scores of 2-3 to reflect proteins that are sequenced but under-characterized, precisely the cases where a structure prediction tool provides the most marginal value over sequence search alone. The total data set had 314 unique proteins distributed evenly across species and split 80%/20% into training and test groups.

### Ablation Experiments

We designed ten ablation experiments (Table 2 and Figure 1) as a broad, but not exhaustive, representation of the various approaches to driving an autonomous workflow. These included a combination of two federation topologies: "monolithic", where all tools were available on a single instance, or "federated", where tools were distributed across individual servers); two experimental harnesses (classic RL-PPO training [Schulman et al., 2017], or prompt-driven LLM harness), and two prompt configurations (one with simple instructions, and one augmented, Table 3). We also explored the performance of the LLM harness using a low-cost or frontier model (o4-mini from OpenAI vs Claude Opus 4.8 from Anthropic). For comparison across methods, and to facilitate the calculation of prediction stability for individual proteins, we ran all experiments against the test dataset with five rollouts.

### Topology handling

Experiments were deployed across three EC2 instances running independent ADEPT stacks, each with its own Keycloak identity provider, MCP tool server, and orchestration service. In the monolithic topology (E1-E3), all tools resolved locally within a single Docker network. In the federated topology (E4-E6), Lab A hosts BLAST, Lab B hosts AlphaFold, and Lab C serves as the orchestrating gateway, with authentication across "Labs" established through service client credentials via KeyCloak. For federated LLM-harness experiments (E5/E6), tools are registered in Lab C's Redis as named remote MCP endpoints; for PPO experiments (E4), the executor calls peer MCP servers directly via JSON-RPC, bypassing the LLM-oriented bridge while still traversing the topology.

### PPO RL specifications (Experiments E1, E4)

The PPO agent received no natural language input. Its observation space was a 9-dimensional continuous vector encoding sequence length (normalized), BLAST confidence, AlphaFold confidence, accumulated cost (normalized against a budget ceiling), and binary indicators for which tools have been called. The action space was Discrete(3): call BLAST, call AlphaFold, or STOP. Episodes can terminate upon STOP or after a maximum of 5 steps. Training used stable-baselines3 PPO with default hyperparameters over 500 episodes on a 120-protein training split spanning all species in the benchmark. No species-specific tuning was applied. Training converged within approximately 15 minutes on a single EC2 instance (no GPU required). We trained three independent seeds per topology condition and report test-set performance only, evaluated on 30 held-out proteins unseen during training by either the PPO agent or the LLM. The LLM-harness experiments (E5/E6) used the same proteins and reward function but undergo no formal policy optimization. Each episode was a single-pass inference through ADEPT's Responses API where the LLM's tool-calling behavior was shaped only by the prompt. Scores were computed post-hoc for comparison, and no gradient signal flows back to the LLM.

The PPO agent's tool calls returned structured numeric features directly from cached responses: BLAST identity percentage (converted to “confidence”, e-value, hit count, AlphaFold pLDDT, without any natural language interpretation. This represents a lower bound on reasoning sophistication: the agent learns purely from reward signal which tool sequences yield favorable cost-accuracy tradeoffs under different latency regimes.

**Reward and cost structure**

The following equation describes the reward function used for the RL-PPO experiments. The components of this equation were also recorded for the LLM-driven experiments, but were not used to train behavior.

$$R = Accuracy - P_{tool\ cost} - P_{latency} + B_{structural} + B_{confidence}$$

Where accuracy is a fixed positive reward for correct predictions and a penalty for incorrect ones, described in more detail below. Cost penalties ($P_{tool\ cost}$, $P_{latency}$) sums per-tool pseudo-dollar costs ($5 BLAST, $10 AlphaFold) and wall clock time, respectively. To provide gradient signal encouraging higher-quality tool outputs, we included structural bonus (weighted pLDDT when AlphaFold is called, $B_{structural}$) and confidence bonus (weighted, normalized BLAST percent identity, $B_{confidence}$).

Accuracy is evaluated via Gene Ontology term matching: predicted GO terms (inferred by annotation transfer from the BLAST top hit's UniProt record) are compared against ground truth GO terms for Molecular Function (F) and Biological Process (P) categories [20]. Critically, only successful use of the BLAST tool contributed to accuracy. AlphaFold provides no route to GO term prediction as it offers structural confidence that shapes reward but cannot transfer functional annotations[2]. A prediction is correct if *any* predicted term overlaps with the ground truth set. We explicitly do not traverse the GO hierarchy or match parent/child terms. GO:0016787 (hydrolase activity) does not satisfy a ground truth of GO:0004185 (serine-type carboxypeptidase activity) even though the latter is a descendant. This presents a concrete, verifiable metric that avoids the asymmetric penalty of exact-match scoring (wherein proteins with more GO terms would be easier to match, distorting learned policies).

**LLM specifications (Experiments E2, E3, E5, E6)**

LLM experiments were, accessed programmatically through the ADEPT orchestration service (ordinarily a Streamlit interface) at temperature 0.0; variation in responses arises via greedy decoding. Each protein constituted a single inference call with no cross-episode memory; "episode" therefore denotes one isolated characterization task rather than a step in a sequential trajectory. Two prompt strategies were compared (Table 3): a minimal zero-shot prompt providing only tool names and a augmented prompt supplying explicit decision thresholds and a four-step routing strategy. Ground-truth BLAST confidence (percent identity scaled to [0,1]) and true tool invocations were drawn from cached benchmark data, while reported confidence and routing decisions were parsed from structured ROUTING_SUMMARY blocks in the model response and from Langfuse execution traces. Prompt templates were refined on the training split solely to ensure consistent ROUTING_SUMMARY formatting and complete trace capture; no model weights were modified.

[2] This reflects biological reality: homology-based annotation transfer (BLAST) is the standard mechanism for functional inference, while structure prediction supports but does not independently determine function.

### Performance metrics

We developed a suite of performance metrics to compare experiments' ($m$) performance for individual proteins ($p$). The suite includes and extends the metrics represented in the cost function for the PPO-RL algorithm (shown above), with the goal of illustrating the tradeoffs associated with various experimental harnesses and federation topologies. Metrics were calculated for 5 rollouts of the test data and summarized for each of the ten experiments (Table 2). The metrics reported were as follows:

1) *Accuracy*: as above, accuracy is obtained when at least one of the predicted GO terms is present in the ground-truth GO dataset. For an individual protein, this is a binary outcome (0=inaccurate, 1 = accurate); for summarization we calculate the sum of accurate predictions divided by the number of total predictions.
2) *Consistency*: This metric represents the per-protein agreement rate within a given experiment,
$$consistency_{p,m} = \max(k, n-k)/n$$
Where k = the number of accurate responses out of $n$ trials for protein $p$ (i.e., the number of rollouts, 5 in our case). This gives a value in [0.5,1.0] where 0.5 means maximally inconsistent and 1.0 means perfectly consistent. Therefore, lower values of this metric indicate approaches that produce varying results across trials, regardless of accuracy.
3) *Tool Efficiency*: This metric captures the quality gained per tool call, to characterize agents that achieve high confidence and accuracy without unnecessary tool invocations.
$$tool\ efficiency_{p,m} = \frac{accuracy_{p,m} + confidence_{p,m}}{2 * cost\ ratio}$$
Where accuracy is as defined above; confidence is the average of the reported BLAST confidence and AlphaFold PLDDT if both tools called (normalized between 0 and 1); and cost ratio is the sum of the tool costs divided by total tool cost possible.
4) *Token use and estimated inference cost*: We obtained the total number of tokens consumed by each trial (including ingress and egress) reported by the Langfuse traces, subtracted the tokens associated with the system prompt (estimated to be 11,000). Estimated inference cost was calculated using average model-specific rates as of July 2026 ($3.43/1m tokens for o4-mini, $18.97/1m tokens for Opus 4.8).
1) *Speed*: The total wall clock time (seconds) taken to complete a trial. These are inverted on Figure 2 so that lower clock times are represented as higher speeds.

We present results across several key axes: the federation topologies and harnesses described in Table 2 as well as performance across proteins with low BLAST confidence (<0.7) or high BLAST confidence (>0.7). This allowed us to evaluate whether performance varies meaningfully for lesser-known proteins, providing insight into how our tested methods might facilitate or hinder autonomous discovery.

## Results

This section presents our results organized by the performance metrics described above, and describes the learned (E1/E4) or emergent (E2, E3, E5, E6) policies from each harness. For each metric, we report the relative performance and highlight key distinctions across experiments and BLAST confidence levels. A high-level summary of experimental performance is presented in Figure 2 and specific values with 90% confidence intervals are shown in Table 4 and Figure 2.

*Learned vs. Emergent Routing Strategies:*

The PPO policy (E1/E4) learned a near-universal alphaFold escalation strategy, invoking the structural tool at high probability across virtually all BLAST confidence and sequence length bins (Figure 3). Most LLM harnesses exhibited the same "always escalate" strategy regardless of BLAST confidence, achieving high accuracy through exhaustive tool use at maximum cost. The notable exception was expert-prompted Opus (E3/E6), which emerged with the opposite behavior (rarely calling AlphaFold) yet maintained >92% accuracy. The poorly-performing outlier, zero-shot o4-mini (E2), also showed low invocation and yielded the worst accuracy overall.

*Accuracy*: Accuracy was higher for all harnesses and both languages under the monolithic topology (ranging from 50.5% for E3 with o4-mini to 93.8 for E2 with Opus 4.8, Table 4 and Figure 2). Accuracy did not change across topologies for the RL harness (87.6%), but degraded slightly for all harnesses and both languages in the federated topology. The LLM harnesses that used Opus 4.8 were the most accurate for both topologies, though the zero-shot prompt slightly outperformed the expert prompt in the monolithic topology (E2 at 93.8% and E3 at 92.5%), a result that reversed for the federated topology (E6 at 92.1% and E5 at 91.5%). The least accuracy across topologies was obtained by the LLM harness using o4-mini (46.6% for E2 [monolithic], and 39.8% for E5 [federated, the worst performing overall]). For all experiments, the 90% CI across the test set encompassed the entire range (0%-100%).

*Consistency:* Consistency was highest for the RL harness in both topologies at 1.000, with a 90% CI of 1.000-1.000, indicating that there was not a single test protein for which the RL harness provided different answers across five rollout trials (Table 4 and Figure 2). The experiments using Opus 4.8 were close behind, with consistency scores that correlated well with accuracy: the most-consistent LLM-driven experiment in the monolithic topology was also the most accurate (E2 at 0.974), whereas the zero-shot prompt with Opus 4.8 remained the most consistent experiment in the federated topology (E5 at 0.980) where it was the second-most accurate experiment. For both topologies, the augmented prompt with Opus 4.8 provided similar levels of consistency (0.977 for both E3 and E6). All Opus 4.8 experiments had 90% CIs for consistency between 0.800 and 1.000. The least consistent experiments were the zero-shot prompts with o4-mini (0.879 for E2 and 0.843 for E5), with a 90% CI between 0.6 and 1.0.

*Tool Efficiency:* Tool efficiency was maximized at similar values by the expert prompting strategy in both topologies (ranging from 0.652 in E3 with o4-mini to 0.686 with Opus 4.8 in E6,

Table 4 and Figure 2). However, the least-tool-efficient experiments for both topologies were those that used the zero-shot prompt with o4-mini, which minimized at a score of 0.000 in the monolithic setting. The RL harness produced a tool efficiency rate of 0.788 which did not vary across topologies.

*Speed:* The RL algorithms were the fastest, taking an average of 15-16 seconds per protein for both topologies, while the LLM harnesses ranged from 36 to 58 seconds (E2 and E6 with o4-mini, respectively, Table 4 and Figure 2). Latency was more variable across topologies for the o4-mini experiments. (E2 was 58s vs. E5 at 41s, while E3 was 46s vs. E6 at 36s). Opus was more stable (~39-42s across all conditions).

*Token Usage and Cost:* The RL-driven experiments consumed no LLM tokens and incurred no inference cost (Table 4 and Figure 2). Among LLM-driven experiments, token consumption diverged sharply by model. The o4-mini experiments consumed a median of ~28,500 tokens per protein under expert prompting (28,554 for E3; 28,297 for E6) and under federated zero-shot prompting (28,860 for E5), but inflated to 45,849 tokens in the monolithic zero-shot condition (E2). Opus token consumption was higher overall and more stable across conditions, ranging from 33,334 (E2, monolithic zero-shot) to 49,136 (E6, federated expert), with expert-prompted experiments consistently consuming more tokens (~48-49k) than zero-shot experiments (~33k) regardless of topology. Per-protein inference cost reflected token volume scaled by each model's pricing. The o4-mini experiments averaged $0.097-$0.099 per protein under expert prompting (E3: $0.0979; E6: $0.0971) and $0.099 under federated zero-shot prompting (E5: $0.0990), with the monolithic zero-shot anomaly reaching $0.1573 (E2). Opus was 6-10x more expensive, ranging from $0.6323 (E2, monolithic zero-shot) to $0.9321 (E6, federated expert), with expert-prompted conditions consistently costlier than zero-shot conditions due to the higher token volume associated with structured tool-use reasoning. Cost did not vary meaningfully across topologies for either model when prompt strategy was held constant (e.g., E3 Opus at $0.9177 vs. E6 Opus at $0.9321).

*Performance across topologies:* Within the LLM-driven harnesses, federation introduced a modest accuracy penalty for all model-prompt combinations: o4-mini zero-shot dropped from 46.6% to 39.8% (-6.8 pp), Opus zero-shot from 93.8% to 91.5% (-2.3 pp), o4-mini expert from 50.5% to 49.2% (-1.3 pp), and Opus expert from 92.5% to 92.1% (-0.4 pp, Table 4 and Figure 2). Expert routing partially recovered the federation penalty: the accuracy gap between monolithic and federated was smallest for the expert-prompted conditions (-1.3 pp and -0.4 pp) compared to zero-shot (-6.8 pp and -2.3 pp).

*Across LLMs:* For the two LLMs examined, Opus consistently dominated o4-mini on accuracy (91.5-93.8% vs. 39.8-50.5%) and consistency (0.974-0.980 vs. 0.843-0.911), at a cost premium of approximately 6-10x ($0.63-$0.93 vs. $0.10-$0.16 per protein, Table 4 and Figure 2). However, o4-mini was faster than Opus under expert prompting (36-46s vs. 39-40s) and consumed fewer tokens in most conditions.

*Across BLAST confidence levels:* The radar plots stratified by BLAST confidence reveal that the RL harness's learned policy was most effective for high-confidence proteins (>0.7), where E1/E4's accuracy polygon expands to nearly match or exceed Opus, suggesting the PPO policy learned to correctly trust BLAST's initial annotation when confidence was already high, economizing on additional tool calls (Figure 2). In the low-confidence regime (≤0.7), the RL harness's accuracy visibly contracts relative to Opus.

## Discussion

This study uses a routine, verifiable life-sciences workflow to examine the behavior of a co-scientist like framework across experimental harnesses and federation topologies. We do not present nor explore a frontier RL-LLM hybrid, since this was not necessary to reach strong performance on benchmarking tasks

We find that for our verifiable, routine workflows, the model's capability is the dominant performance lever, whereas architecture did not have meaningful impacts. The accuracy gap between LLMs (Opus ~92-94% vs. o4-mini ~40-50%) dwarfed the gaps between prompting strategy and between topologies. Introducing federation, including the need for A2A communication, authentication, and trace logging across instances, did not impose a perceived tax on any harness. The RL harness was invariant across topologies (+1 s, identical accuracy), and the o4-mini experiments were actually faster under federation (e.g., 58s vs 41s for zero-shot runs), possibly because the structured A2A delegation constrained reasoning length; this would also explain the discrepancies in token usage across topologies for that LLM. This suggests that architects should first select the inference model to use in their system, given their own cost constraints, and not anticipate a measurable performance or latency penalty.

We also note that learned and emergent policies from frontier LLMs appear to converge on the same routing logic, with clear tradeoffs of observability for consistency and cost. Both a reward-trained PPO policy and expert-prompted Opus independently arrived at the same confidence-conditional strategy: trust BLAST when confidence is high, escalate to AlphaFold when it's low. The diversity bonus (rewarding exploration of AlphaFold as a supplementary tool) appeared less influential than what was already present in the training data. This is aligned with most human researchers' approach to this workflow.

The observability enabled by the LLM harness provides both reasoning and flexibility, and the added opportunity to recover the correct or enriched answer purely from its training data, regardless of tool success. This "opportunity" is of stochastic value, with inconsistency especially pronounced for low-confidence cases, is accompanied by nontrivial token usage (costs) and latency, and is strongly model-dependent. The o4-mini failure mode is particularly illustrative (Table 5). It often reasoned correctly but returned the wrong final answer, an instruction-following/output failure rather than a reasoning one, and in zero-shot it failed to engage tools at all (tool efficiency 0.000). Expert prompting is the highest-leverage intervention for the hardest tasks (our low-confidence proteins) for this weakest model. In contrast, the RL-PPO harness provided perfect consistency (1.000), consumed zero tokens, was fast and robust to topologies, but was wholly rigid, provided no reasoning trace and required a curated training set and verifiable reward. Therefore, we find that for such a repetitive, verifiable workflow, an inexpensive learned policy can match a frontier LLM at a fraction of the cost while providing perfect reproducibility. We recommend that scientists implement expensive, stochastic LLM harnesses for use cases that genuinely need flexible reasoning.

The workflow examined for this study was recovery-only (of a protein's known functional annotation). Real co-scientist work blends recovery (consulting known information) with discovery (conjecture and inference over the unknown. The scientific-agent literature centers this distinction. In our findings, even on well-designed proteins the weakest models (e.g., o4-mini) failed, meaning that naïve autonomy presents risks even for these easy, routine cases that clearly exist in frontier models' training data (Table 5). Conversely, the stronger LLM could perform well on recovery but with unsatisfying consistency (Table 4), which becomes a more pronounced risk when answers are not verifiable (as is the case in all discovery-focused work).

What is the value of the observability suite? For recovery policies like ours, the Langfuse traces do not provide much insight beyond the witnessing of model confabulation and tool call success, which are unlikely to be of much value unless there is a verifier-like agent that consumes them. In contrast, for discovery such reasoning traces are the essential product, yet we found that regime (or proxy thereof) to be exactly where consistency was worst and cost was highest. This indicates that the value of scientific agents' observability suites scale with novelty, and not with autonomy *per se.*

All of the aforementioned tradeoffs are exactly the reason why fine-tuned hybrid models are being developed (to train the LLM toward the verified answer, and avoid trading accuracy for consistency). We suggest that before researchers begin to implement LLM-driven for research tasks, they locate their workflow on a spectrum between "verifiable recovery" and "novel discovery". The closer to our case study (routine, easily verified, recovery task), the more strongly they should consider cheap deterministic policies with light monitoring. For discovery-heavy pipelines, our findings strongly encourage budgeting for a highly capable model, anticipate less-than-perfect consistency, and prioritize observability proportional to the novelty of the desired outcome.

**Acknowledgements**
The authors would like to acknowledge the input of two internal reviewers at the Pacific Northwest National Laboratory.

**Funding Statement**
This research was supported by the Generative AI initiative for ADEPT, under the Laboratory Directed Research and Development (LDRD) Program at Pacific Northwest National Laboratory (PNNL). PNNL is a multi-program national laboratory operated for the U.S. Department of Energy (DOE) by Battelle Memorial Institute under Contract No. DE-AC05-76RL01830. This research was performed at the William R. Wiley Environmental Molecular Science Laboratory (EMSL) at Pacific Northwest National Laboratory, a national scientific user facility sponsored by the U.S. Department of Energy's Office of Biological and Environmental Research.

**Conflict of Interest Statement**

The authors have no conflicts to declare.

**Author Contribution Statement.**

MK: Conceptualization, data curation, formal analysis, investigation, methodology, software, visualization, writing – original draft; TB: Conceptualization, writing – review & editing; AJ: Writing – review & editing; PR: Conceptualization, Funding acquisition, methodology, resources, software, supervision, writing – review and editing

Tables

| Species | Description and Rationale | Average BLAST identity | Average Alphafold PLDDT | Number of proteins in training dataset | Number of proteins in test dataset | Total |
|---|---|---|---|---|---|---|
| *Escherichia coli (*K12*)* | Well characterized bacteria common to bioengineering workflows with high identity hits across databases | 1.000 | 91.3 | 60 | 15 | 75 |
| *Pseudomonas putida* | | 0.991 | 92.1 | 63 | 16 | 79 |
| *Naegleria gruberi* | Free-living amoeba with minimal Swiss-Prot homologs | 0.474 | 84.5 | 68 | 17 | 85 |
| *Hydra vulgaris* | Early-branching cnidarian with novel, poorly represented domain architectures | 0.559 | 83.6 | 63 | 16 | 75 |

Table 1. *Data used in study from four species. "Average BLAST identity" is the normalized percent identity score across all proteins used in our study. "Average AlphaFold PLDDT" is the average predicted local distance difference test across all proteins used in our study.*

| **Experiment ID** | **Federation Topology** | **Harness** | **Evaluation rollouts*** | **Rationale** |
|---|---|---|---|---|
| E1-Mono-NN | Monolithic (1 container) | PPO (feedforward NN) | 5 | Baseline: no federation, no LLM (RL only) |
| E2-Mono-LLM-naive | As in E1 | LLM routing (Claude Opus 4.8 *or* o4-mini), simple prompt variant | As in E1 | LLM without federation or complex prompting; typical user interaction |
| E3-Mono-LLM-smart | As in E1 | LLM routing (Claude Opus 4.6 *or* o4-mini), engineered prompt variant | As in E1 | As in E2, but with a more capable LLM |
| E4-Fed3-NN | 3-gateway A2A federation | As in E1 | As in E1 | As in E1, but in federated context – does A2A coordination overhead induce tradeoffs? |
| E5-Fed3-LLM-naive | As in E4 | As in E2 | As in E1 | As in E2, but in federated context. |
| E6-Fed3-LLM-smart | As in E4 | As in E3 | As in E1 | As in E3, but in federated context |

Table 2. *Ablation experiments. *For the RL experiments, this column represents the number of evaluation rollouts across the held-out test set after training. The number of episodes used in training is described below. For the natural language experiments, this similarly represents the number of times each protein in the held-out test set was passed through the prompt injection harness.*

| Zero-Shot Prompt (E2, E5) | Augmented Prompt (E3, E6) | End of prompt (E2, E3, E5, E6) |
|---|---|---|
| *I have a protein I'd like to characterize. The accession is {accession} and the sequence is {sequence}.*<br><br>*Can you help me figure out what this protein does? Please use the available bioinformatics tools to gather some evidence before giving me your answer.* | *You are a bioinformatics agent. Your task is to predict the biological function of the protein sequence below as Gene Ontology (GO) term IDs.*<br><br>*PROTEIN SEQUENCE (this is your query — use this string for BLAST):*<br>*{sequence}*<br><br>*INTERNAL REFERENCE ID (for bookkeeping only — do NOT pass this as a sequence): {accession}*<br><br>*AVAILABLE TOOLS (use ONLY these):*<br>• *perform_blastp_search_biopython — sequence homology search ($5.00, ~5 sec). Parameters: sequence: the full protein sequence string above (NOT the accession ID); query_accession: "{accession}" (cache key only, not passed to NCBI)*<br>• *get_AlphaFold_prediction_and_store — structure prediction confidence ($10.00, ~10 sec). Parameters: accession_id: "{accession}"*<br><br>*DECISION STRATEGY:*<br>*1. Always start with BLAST. Pass the full sequence string as "sequence". Do not pass "{accession}" as the sequence — it is not a sequence.*<br>*2. If BLAST returns a hit with >= 70% identity, transfer GO annotations from the top hit's title and accession as your prediction. Stop here unless confidence is low.*<br>*3. If BLAST identity is < 70% or returns no hits, call AlphaFold to assess structural confidence. Use pLDDT (reported as globalMetricValue / 100) to support or qualify your function inference.*<br>*4. Predict 1-5 GO term IDs for Molecular Function (F) or Biological Process (P). Format: GO:NNNNNNN (e.g., GO:0016787, GO:0006355)* | *End your response with EXACTLY this block:*<br><br>*ROUTING_SUMMARY:*<br>*tools_used: [list only tools you actually called: blast, AlphaFold, or both]*<br>*predicted_go_terms: [GO:NNNNNNN, GO:NNNNNNN, ...]*<br>*confidence: [0.0-1.0 — based on what you observed in the tool outputs, not a random guess]*<br>*reasoning: [2-3 sentences: what evidence supports your prediction? If you used*<br>*both tools, explain why BLAST alone was insufficient.]* |

Table 3 *Prompts used in LLM-based experimental harnesses.*

**Performance Summary (Overall)**

| | Accuracy | Consistency | Tool Efficiency | Speed | Token Usage Per Protein | Est. Inference Cost per Protein | |
|---|---|---|---|---|---|---|---|
| E1: Solo-RL | 87.6% [0.0%,100.0%] | 1.000 [1.000,1.000] | 0.788 [0.312,0.991] | 15 s [15,15] | No tokens used | N/A | Monolithic Experiments |
| E2: Zeroshot (o4mini) | 46.6% [0.0%,100.0%] | 0.879 [0.600,1.000] | 0.000 [0.000,1.500] | 58 s [26,113] | 45,849 [25,801,86,628] | $0.1573 [$0.0885,$0.2971] | |
| E2: Zeroshot (opus) | 93.8% [0.0%,100.0%] | 0.974 [0.800,1.000] | 0.572 [0.354,0.865] | 42 s [38,47] | 33,334 [32,395,35,971] | $0.6323 [$0.6145,$0.6824] | |
| E3: Expert (o4mini) | 50.5% [0.0%,100.0%] | 0.905 [0.600,1.000] | 0.652 [0.291,1.046] | 46 s [36,80] | 28,554 [25,971,43,392] | $0.0979 [$0.0891,$0.1488] | |
| E3: Expert (opus) | 92.5% [0.0%,100.0%] | 0.977 [0.800,1.000] | 0.686 [0.342,2.700] | 39 s [21,44] | 48,378 [30,877,51,371] | $0.9177 [$0.5857,$0.9745] | |
| E4: Fed-RL | 87.6% [0.0%,100.0%] | 1.000 [1.000,1.000] | 0.788 [0.312,0.991] | 16 s [16,16] | No tokens used | N/A | Federated Experiments |
| E5: Fed-Zeroshot (o4mini) | 39.8% [0.0%,100.0%] | 0.843 [0.600,1.000] | 0.487 [0.289,0.820] | 41 s [31,73] | 28,860 [27,082,37,010] | $0.0990 [$0.0929,$0.1269] | |
| E5: Fed-Zeroshot (opus) | 91.5% [0.0%,100.0%] | 0.980 [0.800,1.000] | 0.572 [0.337,0.866] | 41 s [37,46] | 33,717 [32,756,35,784] | $0.6396 [$0.6214,$0.6788] | |
| E6: Fed-Expert (o4mini) | 49.2% [0.0%,100.0%] | 0.911 [0.600,1.000] | 0.670 [0.291,2.500] | 36 s [19,60] | 28,297 [17,881,31,154] | $0.0971 [$0.0613,$0.1069] | |
| E6: Fed-Expert (opus) | 92.1% [0.0%,100.0%] | 0.977 [0.800,1.000] | 0.686 [0.353,2.700] | 40 s [22,44] | 49,136 [31,365,51,959] | $0.9321 [$0.5950,$0.9857] | |

Table 4. *Summary of performance (median, [90% CI]) for individual experiments (rows). For each performance metric (columns), the shading corresponds to the relative performance of experiment within the monolithic experiments (top section) or federated experiments (bottom section), where darker shading = better performance.*

| **Accession**: P0A746 (*E. coli* strain K12)<br>**Principal function(s):** Iron ion binding, and Peptide-methionine (R)-S-oxide reductase activity, also known as MsrB<br>**GO Terms in UniProt:** 0005506, 0006979, 0008270, 0030091, 0033743, 0033745, 0046686 | | | | | |
|---|---|---|---|---|---|
| | **Trial 1 – Correct** | **Trial 2 – Incorrect** | **Trial 3 – Correct** | **Trial 4 – Correct** | **Trial 5 – Incorrect** |
| **Relevant text from LLM response** | "UniProt unambiguously annotates this protein as MsrB ... the AlphaFold model ...methionine sulfoxide reduction..." | "UniProt annotation directly assigns peptide-methionine sulfoxide reductase activity..." | "Retrieved a curated UniProt entry identifying MsrB activity (peptidyl-methionine (R)-S-oxide reductase)...GO assignments for reductase activity, metal ion binding, and general oxidoreductase are well supported." | "This protein matches MsrB in UniProt...peptide-methionine (R)-S-oxide reductase activity (GO:0003756)..." | "“UniProt clearly annotates this protein as MsrB with methionine sulfoxide reductase activity..." |
| **Predicted GO terms** | **0006979**, 0016682, 0051032 | 0004743, 0005737, 0016682, 0046872 | **0008270**, 0016491, 0051536 | 0003756, **0006979**, 0055114 | 0004656, 0016209, 0051213 |
| **Tools called** | BLAST and AlphaFold | Alphafold | none | none | BLAST |
| **Synthesis** | The response text is correct, and 0006979 appears in the benchmark data. The response mentions the UniProt annotation for the protein despite not calling that tool specifically, likely meaning that this information is in the model's training data. | The response text is again correct, and mentions UniProt, but none of the predicted GO terms match the ground truth. | The LLM presented generally correct information in the text and, despite calling no tools, produced one matching GO term (0008270). However, the term "peptidyl-methionine" is not a synonym for the enzyme itself but rather the product involved in the chemical reaction. | As in Trial 3, but distinct GO terms were produced with one match (0006979). | As in Trial 2, but distinct GO terms were returned and none match the ground truth. |

Table 5. *Characteristic examples from E2 (zero-shot prompt) of harness performance across five trials (columns) of the same protein (top row) using o4-mini. The natural language component of the response is correct for all trials, but . Because only one match within an answer is required for accuracy, multiple versions of the answer are marked as correct, but neither approach provides consistent responses nor performance.*

Figures

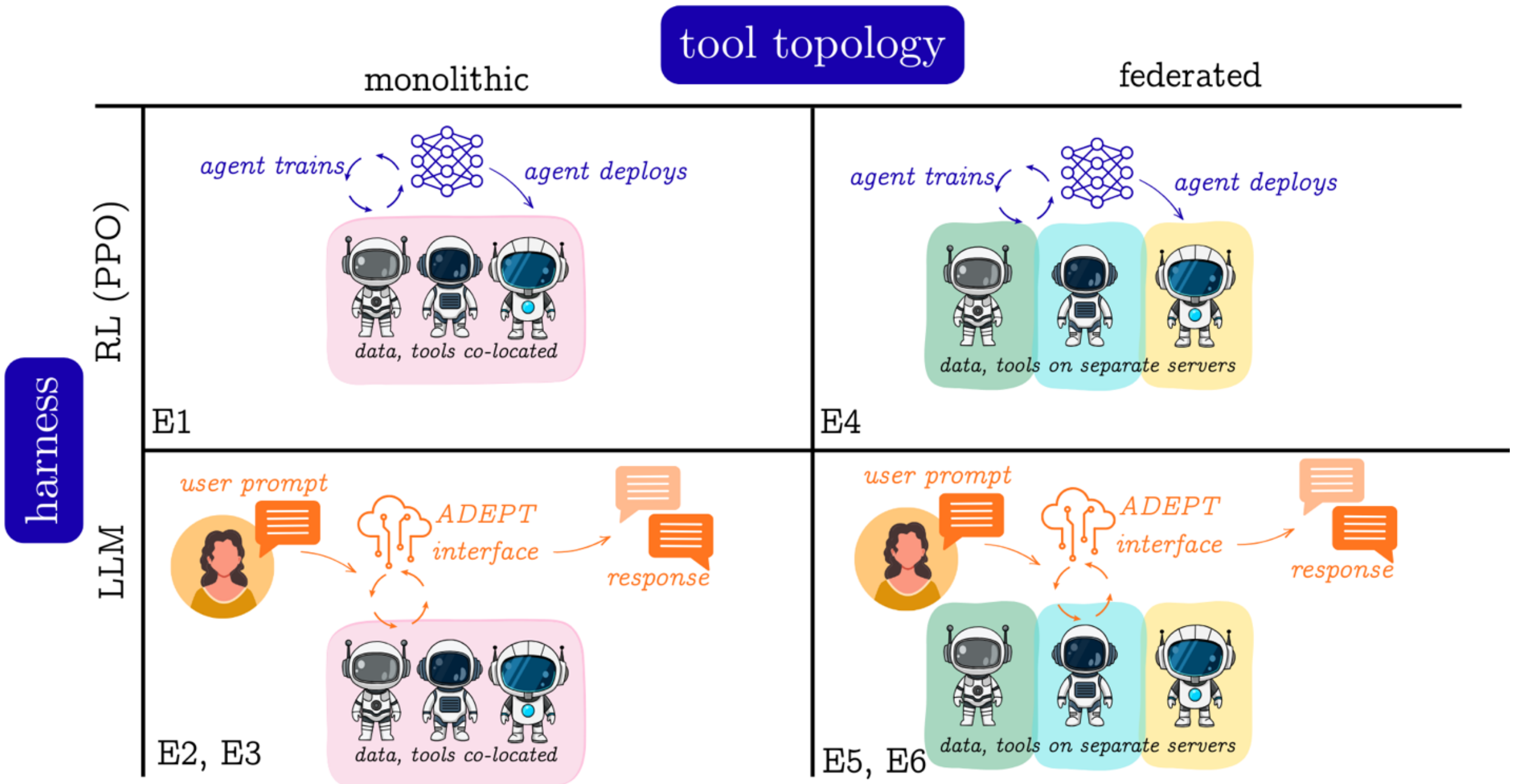

Figure 1 Schematic of study design across two topologies (columns) and experimental harnesses (rows). Corresponding experiments are labeled in the bottom-left of each cell.

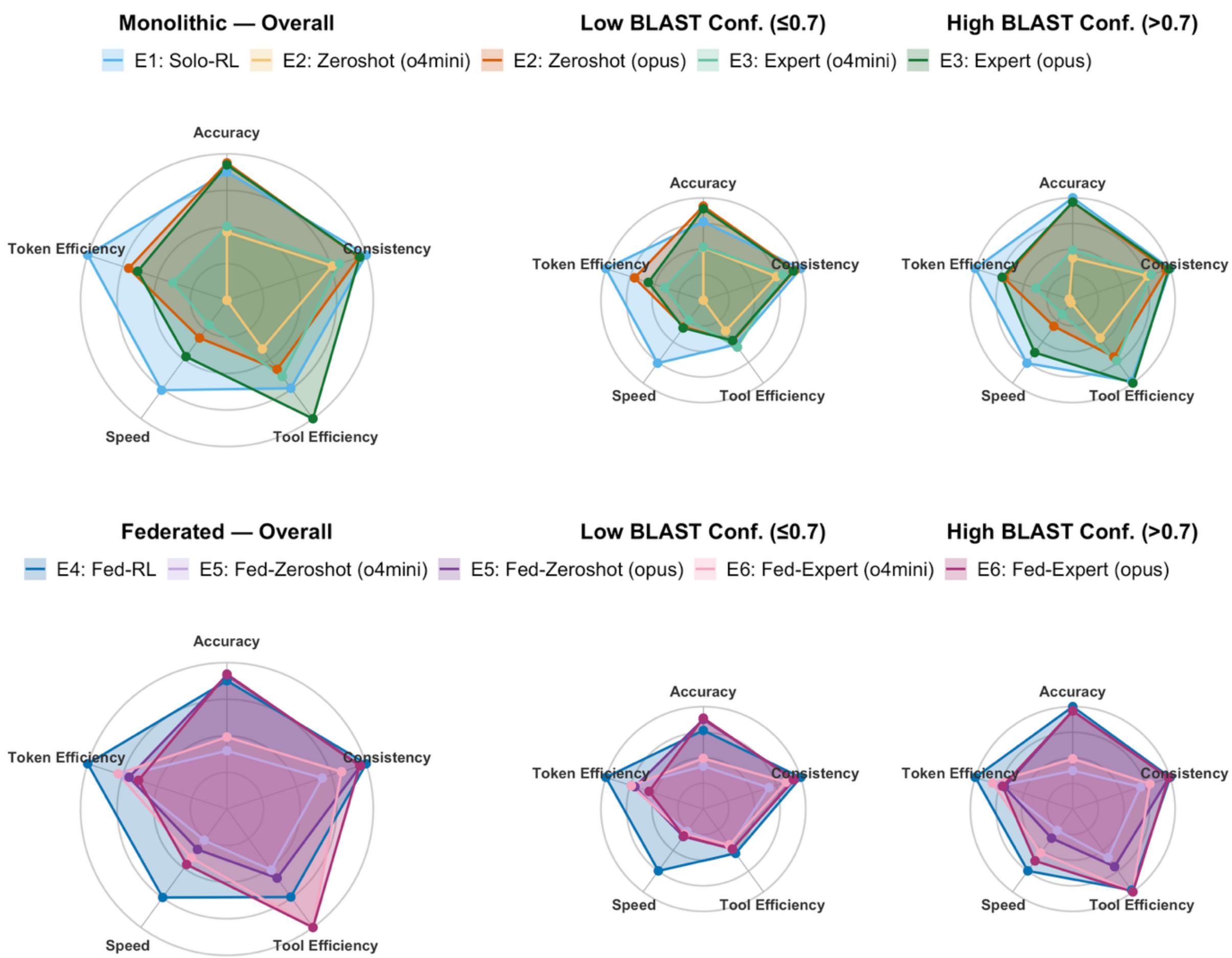

*Figure 2. Performance metrics (labels) across federation topologies (rows) and experimental harnesses (colors) for low-confidence (middle), high-confidence (right), and all examined proteins (left). Points are the median performance following evaluation of either the trained PPO algorithm or the various prompting strategies against all proteins in the test dataset five times.*

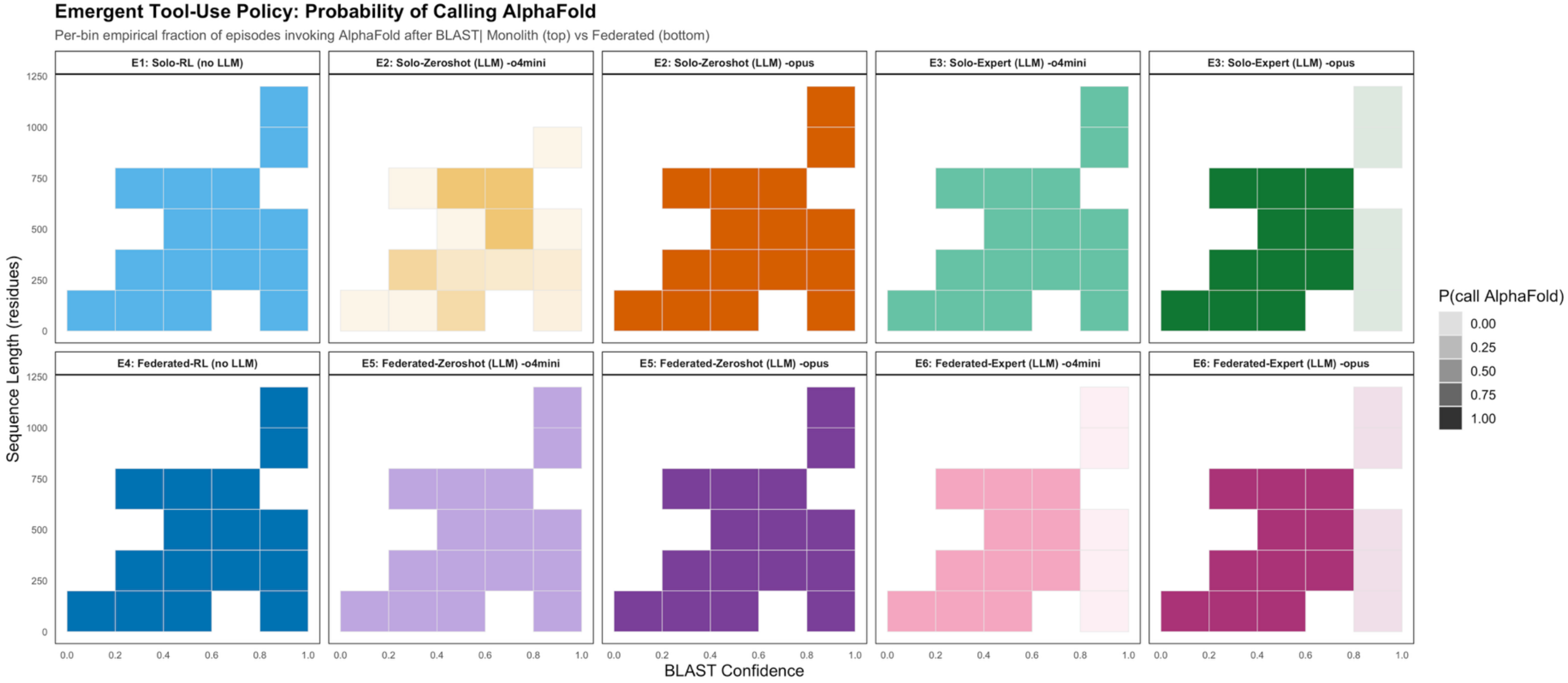

*Figure 3. Illustration of learned policy by PPO RL agent (leftmost column) and emergent policy for LLM driven experiments (second through fifth columns) for all experiments (colors) across topologies (rows) depending upon BLAST confidence level (x axis) or sequence length (y axis).*

Supplementary Material

Figure S1. Overlaid trajectories of learned tradeoffs between normalized tool cost (x axis) and accuracy (y axis) for the two RL experiments (colored labels at endpoints). Trajectories are composed of a 50-episode rolling average of model training across all organisms and confidence levels.

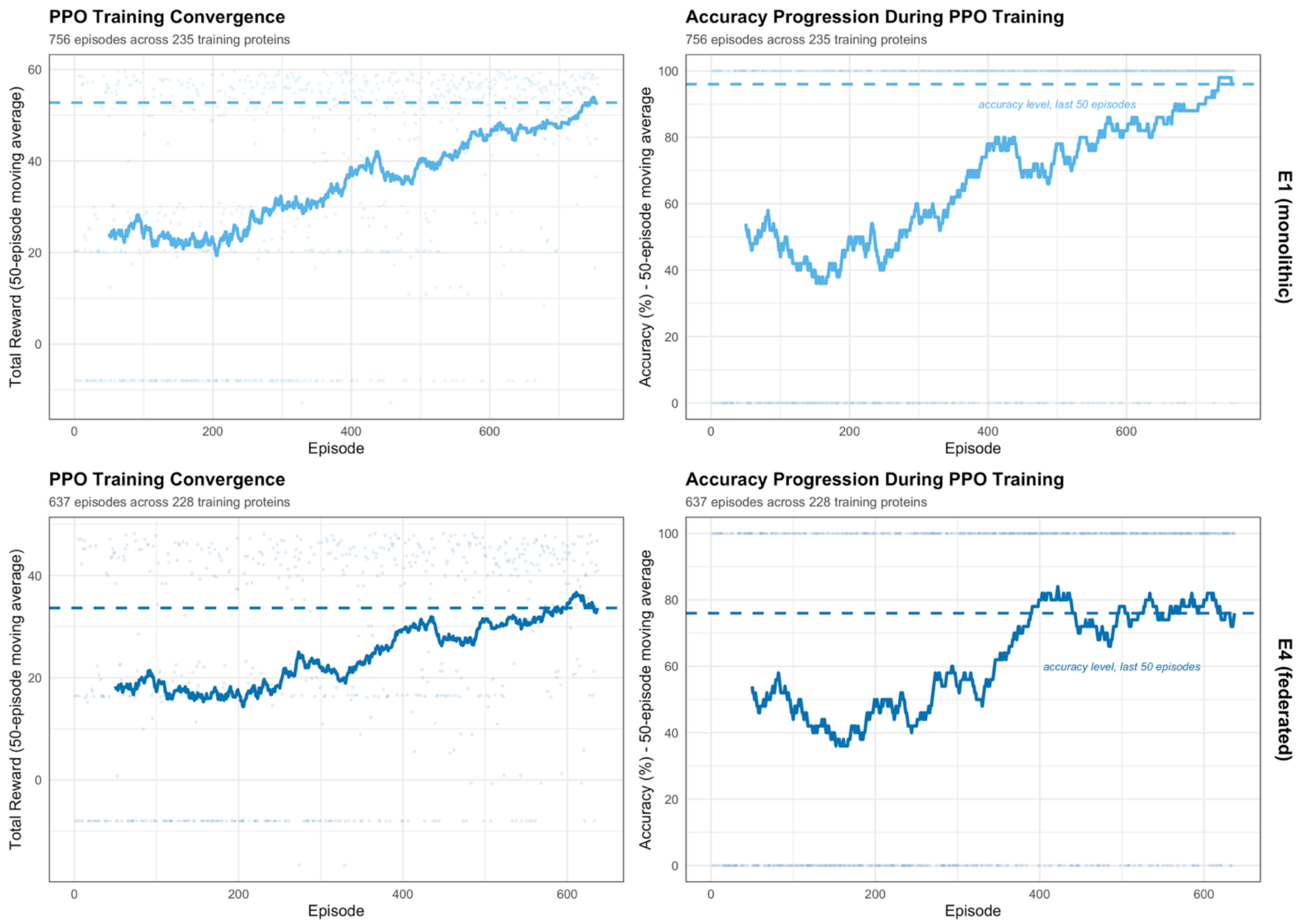

*Figure S2. Training evolution of total reward (left column) and accuracy (right column) across episodes (x axis) for both RL experiments by topology (rows). Each point represents an individual episode, and the dashed line is the average of the terminal 50 episodes.*